# Variational Cumulant Expansions for Intractable Distributions

**David Barber**  DAVIDB@MBFYS.KUN.NL
**Piërre van de Laar**  PIERRE@MBFYS.KUN.NL
*RWCP (Real World Computing Partnership)*
*Theoretical Foundation SNN (Foundation for Neural Networks)*
*University of Nijmegen, CPK1 231, Geert Grooteplein 21*
*Nijmegen, The Netherlands*

## Abstract

Intractable distributions present a common difficulty in inference within the probabilistic knowledge representation framework and variational methods have recently been popular in providing an approximate solution. In this article, we describe a perturbational approach in the form of a cumulant expansion which, to lowest order, recovers the standard Kullback-Leibler variational bound. Higher-order terms describe corrections on the variational approach without incurring much further computational cost. The relationship to other perturbational approaches such as TAP is also elucidated. We demonstrate the method on a particular class of undirected graphical models, Boltzmann machines, for which our simulation results confirm improved accuracy and enhanced stability during learning.

## 1. Introduction

In recent years, interest has steadily grown over many associated fields in representing and processing information in a probabilistic framework. Arguably, the reason for this is that prior knowledge of a problem domain is rarely absolute and the probabilistic framework therefore arises naturally as a candidate for dealing with this uncertainty. Better known examples of models are Belief Networks (Pearl, 1988) and probabilistic neural networks (Bishop, 1995; MacKay, 1995). However, incorporating many prior beliefs can make models of domains so ambitious that dealing with them in a probabilistic framework is intractable, and some form of approximation is inevitable. One well-known approximate technique is Monte Carlo sampling (see e.g., Neal, 1993; Gilks, Richardson, & Spiegelhalter, 1996), which has the benefit of universal applicability. However, not only can sampling be prohibitively slow, but also the lack of a suitable convergence criterion can lead to low confidence in the results. Recently, variational methods have provided a popular alternative, since they not only approximate but can also bound quantities of interest giving, potentially, greater confidence in the results (Jordan, Gharamani, Jaakola, & Saul, 1998; Barber & Wiegerinck, 1999; Barber & Bishop, 1998). The accuracy of variational methods, however, is limited by the flexibility of the variational distribution employed. Whilst, in principle, the flexibility, and therefore, the accuracy of the model, can be increased at will (see e.g., Jaakkola & Jordan, 1998; Lawrence, Bishop, & Jordan, 1998), this generally is at the expense of incurring even greater computational difficulties and optimization problems.

In this article, we describe an alternative approach which is a perturbation around standard variational methods. This has the property of potentially improving the performance at only a small extra computational cost. Since most quantities of interest are derivable from knowledge of the normalizing constant of a distribution, we present an approximation of this quantity which, to lowest order, recovers the standard Kullback-Leibler variational bound. Higher-order corrections in this expansion are expected to improve on the accuracy of the variational solution, despite the loss of a strict bound.

In section (2) we will briefly discuss why the normalizing constant of probability distributions is of such importance. We briefly review the Kullback-Leibler variational bound in section (3) before introducing the perturbational approach in section (4). This approach is illustrated on a well-known class of undirected graphical models, Boltzmann machines, in section (5), in which we also explain the relation of this approach to other perturbational methods such as TAP (Thouless, Anderson, & Palmer, 1977; Kappen & Rodríguez, 1998a). We conclude in section (6) with a discussion of the advantages and drawbacks of this approach compared to better known techniques.

## 2. Normalizing Constants and Generating Functions

We consider a family of probability distributions $Q$ over a vector of random variables $\boldsymbol{s} = \{s_i | i \in [1, \dots, N]\}$, parameterized by $\boldsymbol{w}$,[1]

$$Q(\boldsymbol{s}; \boldsymbol{w}) = \frac{\exp(H(\boldsymbol{s}; \boldsymbol{w}))}{Z(\boldsymbol{w})} . \tag{1}$$

With a slight abuse of notation, we write the normalizing constant as

$$Z(\boldsymbol{w}) = \int d\boldsymbol{s} \, \exp(H(\boldsymbol{s}; \boldsymbol{w})) . \tag{2}$$

The 'potential' $H(\boldsymbol{s}; \boldsymbol{w})$ thus uniquely determines the distribution $Q$. We have assumed here that the random variable $\boldsymbol{s}$ is continuous—if not, the corresponding integral (see equation 2) should be replaced by a summation over all possible discrete states.

One approach in statistics to obtain marginals and other quantities of interest is given by considering generating functions (Grimmett & Stirzaker, 1992), which take the form

$$G(\boldsymbol{\lambda}; \boldsymbol{w}) = \int d\boldsymbol{s} \exp\left(\log Q(\boldsymbol{s}; \boldsymbol{w}) + \boldsymbol{\lambda} \cdot \boldsymbol{s}\right) \tag{3}$$

Averages of variables with respect to $Q$ are given by derivatives of $G(\boldsymbol{\lambda}; \boldsymbol{w})$ with respect to $\boldsymbol{\lambda}$ for fixed $\boldsymbol{w}$, so that, for example, $\langle s_1 s_2 \rangle = \partial^2 G(\boldsymbol{\lambda}; \boldsymbol{w})/\partial \lambda_1 \partial \lambda_2$, evaluated at $\boldsymbol{\lambda} = 0$. We can equally well consider the function

$$Z(\boldsymbol{\lambda}; \boldsymbol{w}) = \int d\boldsymbol{s} \exp\left(H(\boldsymbol{s}; \boldsymbol{w}) + \boldsymbol{\lambda} \cdot \boldsymbol{s}\right) \tag{4}$$

so that

$$\langle s_1 s_2 \rangle = \frac{1}{Z(\lambda_1, \lambda_2; \boldsymbol{w})} \frac{\partial^2 Z(\lambda_1, \lambda_2; \boldsymbol{w})}{\partial \lambda_1 \partial \lambda_2} \bigg|_{\lambda_1, \lambda_2 = 0} . \tag{5}$$

---

1. We assume that the distribution is strictly positive over the domain.

Formally, all quantities of interest can be calculated from "normalizing constants" of distributions, possibly modified in a suitable manner, such as in (4) above. It is clear that all moments of $Q$ can be generated by differentiating (4) in a similar manner, and can be as equally easily obtained from the generating function approach, as from the normalising constant approach. We prefer here the normalising constant approach since this enables us more easily to make contact with results from statistical physics. The normalising constant approach is also more natural for the examples of undirected networks that we shall consider in later sections.

In the following sections, we layout how to approximate the normalising constant, assuming that any additional terms in the normalising constant, of the form $\boldsymbol{\lambda} \cdot \boldsymbol{s}$ in equation (4), are absorbed into the definition of the potential $H$.

As we will see in section (5.2.3), the normalising constant also plays a central role in learning. Unfortunately, for many probability distributions that arise in applications, calculation of the normalizing constant is intractable due to the high dimensionality of the random variable $\boldsymbol{s}$. When $\boldsymbol{s}$ is a $N$–dimensional binary vector, naively at least, a summation over $2^N$ states has to be performed to calculate the normalizing constant.

However, sometimes it is possible to exploit the structure of the distribution in order to reduce the computational expense. A trivial example is that of factorised models, $Q(\boldsymbol{s}) \propto \prod_i \Psi(s_i)$, in which $Z = \int d\boldsymbol{s} \prod_i \Psi(s_i) = \prod_i \int ds_i \, \Psi(s_i)$, so that the computation scales only linearly with $N$. Similarly, in the case of a Gaussian distribution, computation scales (roughly) with $N^3$. These tractable distributions have recently been exploited in the variational method to approximate intractable distributions, see for example (Saul, Jaakkola, & Jordan, 1996; Jaakkola, 1997; Barber & Wiegerinck, 1999; Jordan et al., 1998). In the following section we will briefly review one of the most common variational methods which exploits the Kullback-Leibler bound.

## 3. The Kullback-Leibler Variational Bound

Our aim in this section is to briefly describe the current state-of-the-art in approximating the normalizing constant of an intractable probability distribution. We denote the intractable distribution of interest as $Q_1$ to distinguish it from an auxiliary distribution $Q_0$, we will introduce. Without loss of generality, we write the distribution over a vector of random variables $\boldsymbol{s}$, parameterized by a vector $\boldsymbol{w}$ as

$$Q_1(\boldsymbol{s}; \boldsymbol{w}) = \frac{e^{H_1(\boldsymbol{s};\boldsymbol{w})}}{Z_1(\boldsymbol{w})} \,, \tag{6}$$

where the potential $H_1(\boldsymbol{s}; \boldsymbol{w})$ is a known function of $\boldsymbol{s}$ and $\boldsymbol{w}$, e.g., in the case of a factorised model this would have the form of $H_1 = \sum_k w_k s_k$. The corresponding (assumed intractable) normalizing constant is given by

$$Z_1(\boldsymbol{w}) = \int d\boldsymbol{s} \, e^{H_1(\boldsymbol{s};\boldsymbol{w})} \,. \tag{7}$$

We will use a family of tractable distributions, $Q_0$, parameterized by $\boldsymbol{\theta}$ which we write as

$$Q_0(\boldsymbol{s}; \boldsymbol{\theta}) = \frac{e^{H_0(\boldsymbol{s};\boldsymbol{\theta})}}{Z_0(\boldsymbol{\theta})} \ \text{ with } \ Z_0(\boldsymbol{\theta}) = \int d\boldsymbol{s} \, e^{H_0(\boldsymbol{s};\boldsymbol{\theta})} \tag{8}$$

where, by assumption, the normalizing constant $Z_0$ is tractably computable.

Standard variational methods attempt to find the best approximating distribution $Q_0(\boldsymbol{s}; \boldsymbol{\theta}^*)$ that matches $Q_1(\boldsymbol{s}; \boldsymbol{w})$ by minimizing the Kullback-Leibler divergence,

$$\text{KL}(Q_0, Q_1) = \int d\boldsymbol{s}\, Q_0(\boldsymbol{s}; \boldsymbol{\theta}) \log \frac{Q_0(\boldsymbol{s}; \boldsymbol{\theta})}{Q_1(\boldsymbol{s}; \boldsymbol{w})} \,. \tag{9}$$

Since, using Jensen's inequality, $\text{KL}(Q_0, Q_1) \geq 0$, we immediately obtain the lower bound

$$\log Z_1 \geq -\int d\boldsymbol{s}\, Q_0(\boldsymbol{s}; \boldsymbol{\theta}) \log Q_0(\boldsymbol{s}; \boldsymbol{\theta}) + \int d\boldsymbol{s}\, Q_0(\boldsymbol{s}; \boldsymbol{\theta}) H_1(\boldsymbol{s}; \boldsymbol{w}) \,. \tag{10}$$

Using the definition (8) we obtain the bound in the more intuitive form,

$$\log Z_1 \geq \log Z_0 + \int d\boldsymbol{s}\, Q_0(\boldsymbol{s}; \boldsymbol{\theta}) \left( H_1(\boldsymbol{s}; \boldsymbol{w}) - H_0(\boldsymbol{s}; \boldsymbol{\theta}) \right) \tag{11}$$

The lower bound is made as tight as possible by maximizing the right-hand side with respect to the parameters $\boldsymbol{\theta}$ of $Q_0$, which corresponds to minimising the Kullback-Leibler divergence (9). This approach has recently received attention in the artificial intelligence community and has been demonstrated to be a useful technique (Saul et al., 1996; Jaakkola, 1997; Barber & Wiegerinck, 1999). Note that whilst this lower bound is useful in providing an objective comparison of two approximations to the normalising constant (the approximation with the higher bound value is preferred), it does not translate directly into a bound on conditional probabilities. An upper bound on the normalising constant is also required in this case. Whilst, in some cases, this may be feasible, in general this tends to be a rather more difficult task, and we restrict our attention to the lower bound (Jaakkola, 1997).

In the next section, we describe another approach that enables us to exploit further such tractable distributions.

## 4. The Variational Cumulant Expansion

In order to extend the Kullback-Leibler variational lower bound with a view to improving its accuracy without much greater computational expense, we introduce the following family of probability distributions $Q_\alpha$, parameterized by $\alpha$, $\boldsymbol{w}$ and $\boldsymbol{\theta}$:

$$Q_\alpha(\boldsymbol{s}; \boldsymbol{w}, \boldsymbol{\theta}) = \frac{e^{H_\alpha(\boldsymbol{s}; \boldsymbol{w}, \boldsymbol{\theta})}}{Z_\alpha(\boldsymbol{w}, \boldsymbol{\theta})} \tag{12}$$

where the potential is given by

$$H_\alpha(\boldsymbol{s}; \boldsymbol{w}, \boldsymbol{\theta}) = \alpha H_1(\boldsymbol{s}; \boldsymbol{w}) + (1 - \alpha) H_0(\boldsymbol{s}; \boldsymbol{\theta}) \,. \tag{13}$$

The probability distributions in this family interpolate between a tractable distribution, $Q_0$ which is obtained for $\alpha = 0$, and an intractable distribution, $Q_1$ which is obtained for $\alpha = 1$. For intermediate values of $\alpha \in (0, 1)$, the distributions remain intractable. The normalizing constant of a distribution from this family is given by

$$\begin{aligned} \log Z_\alpha(\boldsymbol{w}, \boldsymbol{\theta}) &= \log \int d\boldsymbol{s}\, e^{H_0(\boldsymbol{s}; \boldsymbol{\theta}) + \alpha\left(H_1(\boldsymbol{s}; \boldsymbol{w}) - H_0(\boldsymbol{s}; \boldsymbol{\theta})\right)} \\ &= \log Z_0 + \log \left\langle e^{\alpha\left(H_1(\boldsymbol{s}; \boldsymbol{w}) - H_0(\boldsymbol{s}; \boldsymbol{\theta})\right)} \right\rangle_0 \,, \end{aligned} \tag{14}$$

where $\langle . \rangle_0$ denotes expectation with respect to the tractable distribution $Q_0$. The final term in (14) is intractable for any $\alpha \neq 0$, and we shall therefore develop a Taylor series expansion for it around $\alpha = 0$. That is,

$$\log \left\langle e^{\alpha \left(H_1(\boldsymbol{s};\boldsymbol{w}) - H_0(\boldsymbol{s};\boldsymbol{\theta})\right)} \right\rangle_0 \approx \alpha \langle H_1(\boldsymbol{s};\boldsymbol{w}) - H_0(\boldsymbol{s}) \rangle_0$$
$$+ \frac{\alpha^2}{2} \left(H_1(\boldsymbol{s};\boldsymbol{w}) - H_0(\boldsymbol{s}) - \langle H_1(\boldsymbol{s};\boldsymbol{w}) \rangle + \langle H_0(\boldsymbol{s}) \rangle_0\right)^2 + O(\alpha^3) \quad (15)$$

In terms of the random variable $\Delta H \equiv H_1(\boldsymbol{s};\boldsymbol{w}) - H_0(\boldsymbol{s};\boldsymbol{\theta})$, we see that (15) contains the first terms in a power series in $\Delta H$, the first being the mean, and the second the variance of $\Delta H$. More formally, we recognize the final term in (14) as the cumulant generating function $K_{\Delta H}(\alpha) \equiv \log \left\langle e^{\alpha \Delta H} \right\rangle_0$, with respect to the random variable $\Delta H$ (Grimmett & Stirzaker, 1992). If the logarithm of the moment generating function $\left\langle e^{\alpha \Delta H} \right\rangle_0$ is finite in the neighborhood of the origin then the cumulant generating function $K_{\Delta H}(\alpha)$ has a convergent Taylor series,

$$K_{\Delta H}(\alpha) = \sum_{n=1}^{\infty} \frac{1}{n!} k_n^0(\Delta H) \alpha^n \ , \quad (16)$$

where $k_n^0(\Delta H)$ is the $n$th cumulant of $\Delta H$ with respect to the distribution $Q_0$. (The definition of the $n$th cumulant is $k_n^0(\Delta H) = \frac{\partial^n}{\partial \alpha^n} \log \left\langle e^{\alpha \Delta H} \right\rangle_0 \big|_{\alpha=0}$). For convenience we have no longer denoted the explicit dependence on the parameters $\boldsymbol{\theta}$ and $\boldsymbol{w}$.

Since the intractable normalizing constant $Z_1$ corresponds to setting $\alpha = 1$ in equation (14), we can write the Taylor series as

$$\log Z_1 = \log Z_0 + \sum_{n=1}^{l} \frac{1}{n!} k_n^0(\Delta H) + \frac{1}{(l+1)!} k_{l+1}^\xi(\Delta H) \ , \quad (17)$$

where the remainder follows from the the Mean Value Theorem (see, for example Spivak (1967)), and $k_{l+1}^\xi(\Delta H)$ is the $(l+1)$th cumulant of $\Delta H$ with respect to $Q_\xi$, $0 \leq \xi \leq 1$. An approximation is obtained by simply neglecting this (intractable) remainder and using the truncated expansion:

$$\log Z_1 \approx \log Z_0 + \sum_{n=1}^{l} \frac{1}{n!} k_n^0(\Delta H) \ , \quad (18)$$

Note that the right-hand side of equation (18) depends on the free parameters $\boldsymbol{\theta}$, and we shall discuss later how these parameters can be set to improve the accuracy of the approximation.

In the following two subsections, we will look in more detail at the first and second-order approximation of this Taylor series, since these illuminate the variational method and the difficulties in retaining a lower bound.

### 4.1 First Order and the Variational Bound

The Taylor series (equation (17)) up to first order yields

$$\log Z_1 = \log Z_0 + \langle \Delta H \rangle_0 + \frac{1}{2} k_2^\xi(\Delta H) \ , \quad (19)$$

where the remainder $k_2^\xi(\Delta H)$ is equal to the variance (the second cumulant) of $\Delta H$ with respect to $Q_\xi$ with $0 \leq \xi \leq 1$, i.e., $k_2^\xi = \text{var}_\xi(\Delta H) = \left\langle \Delta H^2 \right\rangle_\xi - \left\langle \Delta H \right\rangle_\xi^2$, and $\langle . \rangle_\xi$ denotes expectation with respect to the distribution $Q_\xi$. Since the variance is nonnegative for any value of $\xi$, we have

$$\log Z_1 \geq \log Z_0 + \langle \Delta H \rangle_0 , \tag{20}$$

so that $\log Z_1$ is not only approximated but also bounded from below by the right hand side of equation (20). This bound is the standard mean field bound (see also equation (10)), and is employed in many variational methods (see, for example Saul et al., 1996; Jaakkola, 1997; Barber & Wiegerinck, 1999). This bound can then be made as tight as possible by optimizing with respect to the free parameters $\boldsymbol{\theta}$ of the tractable distribution $Q_0$.

### 4.2 Second and Higher Order

To second order equation (17) yields

$$\log Z_1 = \log Z_0 + \langle \Delta H \rangle_0 + \frac{1}{2}\text{var}_0(\Delta H) + \frac{1}{6}k_3^\xi(\Delta H) , \tag{21}$$

with the remainder $k_3^\xi = \left\langle \Delta H^3 \right\rangle_\xi - 3\left\langle \Delta H^2 \right\rangle_\xi \langle \Delta H \rangle_\xi + 2\langle \Delta H \rangle_\xi^3$ and $0 \leq \xi \leq 1$. Since this remainder cannot be bounded in a tractable manner for general distributions $Q_0$ and $Q_1$, we have no obvious criterion to prefer one set of parameters $\boldsymbol{\theta}$ of the tractable distribution over another. Similarly, for higher-order expansions it is unclear, within this expansion, how to obtain a tractable bound and, consequently, how to select a set of parameters $\boldsymbol{\theta}$. We stress that any criterion is essentially a heuristic since the error made by the resulting approximation is, by assumption, not tractably computable. Whilst, in principle, a whole range of criteria could be considered in this framework, we mention here only two. The first, the bound criterion, is arguably the simplest and most natural choice. The second criterion we briefly mention is the independence method, which we discuss mainly for its relation to the TAP method of statistical physics (Thouless et al., 1977).

#### 4.2.1 BOUND CRITERION

The first criterion assigns the free parameters $\boldsymbol{\theta}$ to maximize the (first-order) bound, equation (20). The resulting distribution $Q_0(\boldsymbol{s}; \boldsymbol{\theta}^*)$ is used in the truncated approximation, equation (18). To second order this is explicitly given by

$$\log Z_1 \approx \log Z_0^* + \langle \Delta H \rangle_0^* + \frac{1}{2}\text{var}_0^*(\Delta H) , \tag{22}$$

where a '*' denotes that the distribution $Q_0(\boldsymbol{s}; \boldsymbol{\theta}^*)$ is to be used. Note that the computational cost involved over that of the variational method of section (4.1) is small. No further optimization is required, only the calculation of the second cumulant $\text{var}_0^*(\Delta H)$ which by assumption is tractable.

#### 4.2.2 INDEPENDENCE CRITERION

The second criterion we consider is based on the fact that the exact value of $Z_1$ is independent of the free parameters $\boldsymbol{\theta}$, i.e.,

$$\frac{\partial \log Z_1}{\partial \theta_i} \equiv 0 \ . \tag{23}$$

Since $\log Z_1$ is intractable, we substitute the truncated cumulant expansion into equation (23),

$$\frac{\partial}{\partial \theta_i}\left(\log Z_0 + \sum_{n=1}^{l} \frac{1}{n!} k_n^0(\Delta H)\right) \equiv 0 \ . \tag{24}$$

In general, there are many solutions resulting from the independence criterion and alone it is too weak to provide reliable solutions. Under certain restrictions, however, this approach is closely related to the TAP approach of statistical mechanics (Thouless et al., 1977; Plefka, 1982; Kappen & Rodríguez, 1998a, 1998b), as will also be described in section (5.1.2).

### 5. Boltzmann Machines

To illustrate the foregoing theory, we will consider a class of discrete undirected graphical models, Boltzmann machines (Ackley, Hinton, & Sejnowski, 1985). These have application in artificial intelligence as stochastic connectionist models (Jordan et al., 1998), in image restoration (Geman & Geman, 1984), and in statistical physics (Itzykson & Drouffe, 1989). The potential of a Boltzmann machine, with binary random variables $s_i \in \{0, 1\}$, is given by

$$H(\boldsymbol{s}; \boldsymbol{w}) = \sum_i w_i s_i + \frac{1}{2}\sum_{i,j} w_{ij} s_i s_j \tag{25}$$

with $w_{ij} \equiv w_{ji}$ and $w_{ii} \equiv 0$. Unfortunately, Boltzmann machines are in general intractable since calculation of the normalizing constant involves a summation over an exponential number of states. In an early effort to overcome this intractability, Peterson and Anderson (1987) proposed the variational approximation using a factorised model as a fast alternative to stochastic sampling. However, Galland (1993) pointed out that this approach often fails since it inadequately captures second order statistics of the intractable distribution. Using the potentially more accurate TAP approximation did not overcome these difficulties and often lead to unstable solutions (Galland, 1993). Furthermore, it is not clear how to extend the TAP approach to deal with using more complex, non-factorised approximating distributions. We examine the relationship of our approach to the TAP method in section (5.1.2). Another approach which aims to improve the accuracy of the correlations, though not the normalising constant, is linear response (Parisi, 1988) which was applied to Boltzmann machines by Kappen and Rodríguez (1998a) for the case of using a factorised model, see section (5.1). The approach we take here is a little different in that we wish to find a better approximation primarily to the normalising constant. Approaches such as linear response can then be applied to this approximation to derive approximations for the correlations, if desired.

More flexible approximating distributions have also been considered in the variational approach, for example, mixtures of factorised distributions (Jaakkola & Jordan, 1998; Lawrence et al., 1998), and decimatable structures (Saul & Jordan, 1994; Barber & Wiegerinck, 1999). We show how to combine the use of such more flexible approximating distributions with higher-order corrections to the variational procedure. For clarity and simplicity, we will initially approximate the normalizing constant of a general intractable Boltzmann machine $Q_1$, see equation (25), using only a factorised Boltzmann machine as our tractable model $Q_0$. Subsequently, in our simulations, we will use a more flexible tractable model, a decimatable Boltzmann machine.

### 5.1 Using Factorised Boltzmann Machines

The potential of a factorised Boltzmann machine is given by

$$H_0 = \sum_i \theta_i s_i \,. \tag{26}$$

For this simple model, calculating higher-order cumulants is straightforward once the value of the free parameters $\boldsymbol{\theta}$ are known since,

$$\langle s_{i_1} s_{i_2} \ldots s_{i_l} \rangle_0 = \langle s_{i_1} \rangle_0 \langle s_{i_2} \rangle_0 \ldots \langle s_{i_l} \rangle_0 \text{ with } i_1 \neq i_2 \neq \ldots \neq i_l \,. \tag{27}$$

For notational convenience, we define $m_i$ to be the expectation value of $s_i$,

$$m_i = \langle s_i \rangle_0 = (1 + \exp(-\theta_i))^{-1}. \tag{28}$$

#### 5.1.1 BOUND CRITERION

Using a factorised distribution to approximate the intractable normalizing constant corresponding to equation (25) gives, from equation (20), the variational bound

$$\log Z_1 \geq \mathcal{S}(\boldsymbol{m}) + \sum_i w_i m_i + \frac{1}{2} \sum_{i,j} w_{ij} m_i m_j \,, \tag{29}$$

where, for convenience, we have defined the entropy

$$\mathcal{S}(\boldsymbol{m}) = -\sum_i \{m_i \log m_i + (1 - m_i) \log(1 - m_i)\} \,. \tag{30}$$

One can either maximize the bound, i.e., the right-hand side of equation (29), directly, or use the fact that at the maximum,

$$\frac{\partial}{\partial \theta_k} \left[ \mathcal{S}(\boldsymbol{m}) + \sum_i w_i m_i + \frac{1}{2} \sum_{i,j} w_{ij} m_i m_j \right] = 0 \,, \tag{31}$$

which leads to the fixed point equation (with the understanding that the means $\boldsymbol{m}$ are related to the parameters $\boldsymbol{\theta}$ by equation (28))

$$\theta_i = w_i + \sum_j w_{ij} m_j. \tag{32}$$

VARIATIONAL CUMULANT EXPANSIONS FOR INTRACTABLE DISTRIBUTIONS

The well-known "mean field" equations (Parisi, 1988) are a re-expression of (32) in terms of the means only, given by (28). The optimal parameters of the factorised model might therefore also be obtained by iterating, either synchronously or asynchronously (Peterson & Anderson, 1987), these fixed point equations. Note that equation (31) also holds for minima and saddle-points, so in general additional checks are needed to assert that the parameters correspond to a maximum of (29). Insertion of the fixed point solution into the second-order expansion for the bound criterion, equation (22), yields (up to second order)

$$\log Z_1 \approx \mathcal{S}(\boldsymbol{m}) \sum_i w_i m_i \frac{1}{2} \sum_{i,j} w_{ij} m_i m_j \frac{1}{4} \sum_{i,j} w_{ij}{}^2 m_i(1-m_i) m_j(1-m_j), \tag{33}$$

where we emphasize that the $m_i$ are given by the variational bound solution, i.e., equation (32).

5.1.2 INDEPENDENCE CRITERION

Using a factorised distribution to approximate the intractable normalizing constant, see equation (25), gives up to second order

$$\log Z_1 \approx \mathcal{S}(\boldsymbol{m}) + \sum_i w_i m_i + \frac{1}{2} \sum_{i,j} w_{ij} m_i m_j + \frac{1}{4} \sum_{i,j} w_{ij}{}^2 m_i(1-m_i) m_j(1-m_j)$$
$$+ \frac{1}{2} \sum_i \left( \theta_i - w_i - \sum_j w_{ij} m_j \right)^2 m_i(1-m_i) \ . \tag{34}$$

The independence criterion leads to the equations

$$\left( \sum_j w_{ij}{}^2 m_j(1-m_j) + \left( \theta_i - w_i - \sum_j w_{ij} m_j \right)^2 \right) \left( \frac{1}{2} - m_i \right)$$
$$= \sum_j \left( \theta_j - w_j - \sum_k w_{jk} m_k \right) w_{ij} m_j(1-m_j) \ . \tag{35}$$

In general, there are many solutions to these equations and the search can be limited by inserting in equation (34) the constraint that the second-order solution is close to the first-order solution, i.e., $\theta_i = w_i + \sum_j w_{ij} m_j + \mathcal{O}(w^2)$, and by neglecting terms of $\mathcal{O}(w^4)$ and higher. This simplifies equation (34) to $-F_{\text{TAP}}$, the negative TAP free energy[2] (Thouless et al., 1977; Plefka, 1982; Kappen & Rodríguez, 1998a). The independence criterion, i.e., $\partial F_{\text{TAP}}/\partial \theta = 0$, leads now to the fixed point condition

$$\theta_i = w_i + \sum_j w_{ij} m_j + \frac{1}{2} \sum_j w_{ij}{}^2 (1-2m_i) m_j(1-m_j) \ , \tag{36}$$

These TAP equations can have poor convergence properties and often produce a poor solution (Bray & Moore, 1979; Nemoto & Takayama, 1985; Galland, 1993). The additional constraint that the TAP solution should correspond to a minimum of $F_{\text{TAP}}$ improves the solution. However, since TAP is therefore essentially an expansion around the first-order solution, we expect the numerical difference between the bound criterion and convergent TAP results to be small. For these reasons, we prefer the straightforward bound criterion and leave other criteria for separate study.

---

2. In Thouless et al. (1977), Plefka (1982) and Kappen and Rodríguez (1998a) the random variables $s_i \in \{-1, 1\}$.

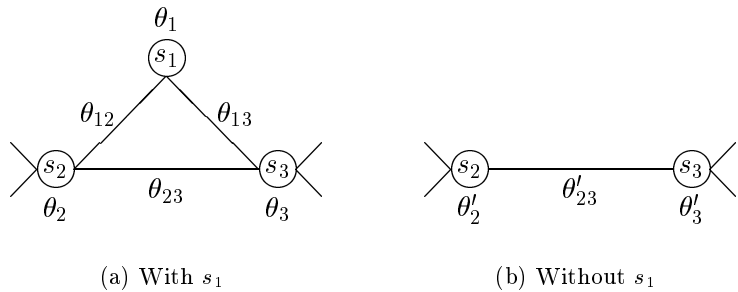

(a) With $s_1$  (b) Without $s_1$

Figure 1: The Boltzmann machine with the random variable $s_1$ can be decimated to a Boltzmann machine without $s_1$.

### 5.2 Numerical Results

In this section we present results of computer simulations to validate our approach. In section (5.2.1) we will compare the different methods of section (4.1) and section (4.2) on approximating the normalizing constant $Z$. A similar comparison is made for approximating the correlations in section (5.2.2). In section (5.2.3) we show the results of a learning problem in which a Boltzmann machine is used to learn a set of sample patterns, a typical machine learning problem. In all cases, the numbers of random variables in the Boltzmann machines are chosen to be small to facilitate comparisons with the exact results.

In some simulations we will not only use factorised Boltzmann machines but also more flexible models that can make the variational bound (see equation (20)) tighter, namely decimatable Boltzmann machines (Saul & Jordan, 1994; Rüger, 1997; Barber & Wiegerinck, 1999). Decimation is a technique that eliminates random variables so that the normalizing constant for the distribution on the remaining variables remains unchanged up to a constant known factor. For completeness, we briefly describe this technique in the current context.

Suppose we have a Boltzmann machine with many random variables, for which a three-variable subgraph is depicted in Figure 1(a), so that random variable $s_1$ is connected only to random variables $s_2$ and $s_3$. Mathematically, the condition for invariance of $Z$ after removing random variable $s_1$, Figure 1(b), becomes

$$\sum_{\boldsymbol{s}\backslash s_1} e^{\theta' + \theta'_2 s_2 + \theta'_3 s_3 + \theta'_{23} s_2 s_3} = \sum_{\boldsymbol{s}} e^{\theta + \theta_1 s_1 + \theta_2 s_2 + \theta_3 s_3 + \theta_{12} s_1 s_2 + \theta_{13} s_1 s_3 + \theta_{23} s_2 s_3}, \qquad (37)$$

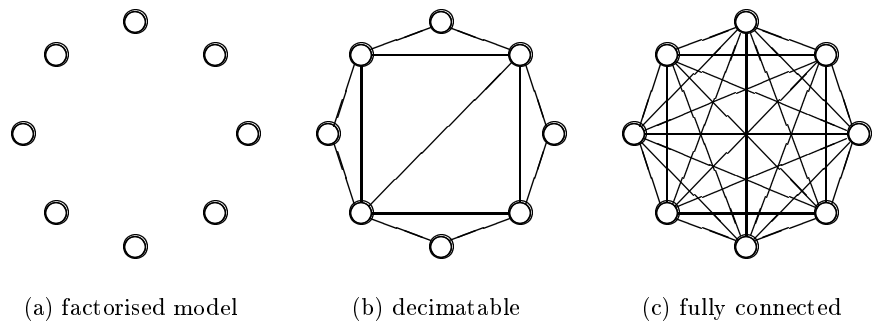

Figure 2: Boltzmann machines of eight random variables

where we include the constant $\theta$ for convenience. Invariance is fulfilled if

$$\theta_2' = \theta_2 + \log \frac{1 + e^{\theta_1+\theta_{12}}}{1 + e^{\theta_1}} \,,\ \theta_3' = \theta_3 + \log \frac{1 + e^{\theta_1+\theta_{13}}}{1 + e^{\theta_1}} \,,$$

$$\theta_{23}' = \theta_{23} + \log \frac{\left(1 + e^{\theta_1}\right)\left(1 + e^{\theta_1+\theta_{12}+\theta_{13}}\right)}{\left(1 + e^{\theta_1+\theta_{12}}\right)\left(1 + e^{\theta_1+\theta_{13}}\right)} \ \text{and}\ \ \theta' = \theta + \log\left(1 + e^{\theta_1}\right) \,. \tag{38}$$

For decimatable Boltzmann machines, one can repeat the decimation process until, finally, one ends up with a Boltzmann machine which consists of a single random variable whose normalizing constant is trivial to compute. This means that for decimatable Boltzmann machines the normalizing constant can be computed in time linear in the number of variables. Decimation in undirected models is similar in spirit to variable elimination schemes in graphical models. For example, in directed belief networks, "bucket elimination" enables variables to be eliminated by passing compensating messages to other buckets (collections of nodes) in such that the desired marginal on the remaining nodes remains the same (Dechter, 1999).

### 5.2.1 Approximating the Normalizing Constant

As our 'intractable' distribution $Q_1$, we took a fully connected, eight-node Boltzmann machine, which is not decimatable (see Figure 2(c)). As our tractable Boltzmann machine, $Q_0$, we took both the factorised model of Figure 2(a) and the decimatable model of Figure 2(b). We then tried to approximate the normalizing constant for the fully connected machine in which the parameters were randomly drawn from a zero-mean, Gaussian distribution with unit variance. See the appendix for additional theoretical and experimental details regarding the optimization scheme used. The relative error of the approximation,

$$E = \frac{\log Z^{\text{exact}} - \log Z^{\text{approx}}}{\log Z^{\text{exact}}} \,, \tag{39}$$

was then determined for both the first and second-order approach using both the factorised and decimatable model. This was repeated 550 times for different random drawings of the parameters of the fully connected Boltzmann machine. The resulting histograms are depicted in Figure 3. In these histograms one can clearly see that the bound, while present in the first-order approach, is lost in the second-order approach since some normalising constant estimates were above the true value, violating the lower bound. In five out of 1100 cases (two times 550) the synchronous mean field iterations did not converge and these outliers were neglected in the plots. It is, however, possible to make the number of non-convergent cases even lower by using asynchronous instead of synchronous updates of the mean field parameters (Peterson & Anderson, 1987; Ansari, Hou, & Yu, 1995; Saul et al., 1996). Note that, in both cases, the second-order method improves on average on the standard (first order) variational results. Indeed, using only a factorised model with the second-order correction improves on the standard variational decimatable result.

We also determined whether the second-order approach improves the accuracy of the prediction in each individual run. In order to determine this we used the paired difference,

$$\Delta = \left|\frac{\log Z^{\text{exact}} - \log Z^{\text{first}}}{\log Z^{\text{exact}}}\right| - \left|\frac{\log Z^{\text{exact}} - \log Z^{\text{second}}}{\log Z^{\text{exact}}}\right|. \tag{40}$$

If $\Delta$ is positive the second order improves the first order approach. The corresponding results of our computer simulations are depicted in Figure 4. In all runs $\Delta$ was positive, corresponding to a consistent improvement in accuracy.

### 5.2.2 Approximating the Means and Correlations

There are many techniques that can be called upon to approximate moments and correlations. Primarily, the cumulant expansion, as we have described it, is intended to improve the accuracy in estimating the normalising constant, and not directly the moments of the distribution. Arguably the simplest way to approximate correlations is to use the standard variational approach to find the best approximating distribution to $Q_1$ and then to approximate the moments of $Q_1$ by the moments of $Q_0$. In this approach, however, certain correlations that are not present in the structure of the approximating distribution $Q_0$, will be trivially approximated. For example, approximating $\langle s_i s_j \rangle_{Q_1}$ using the factorised distribution gives $\langle s_i s_j \rangle_{Q_1} \approx \langle s_i \rangle_{Q_0} \langle s_j \rangle_{Q_0}$. We will examine in this section some ways of improving the estimation of the correlations.

One procedure that can be used to approximate correlations such as $\langle s_i s_j \rangle_{Q_1}$ is given by the so-called linear response theory (Parisi, 1988). This approach uses the relationship (5) in which the intractable normalising constant is replaced with its approximation (29).

The approach that we adopt here, due to it's straightforward relationship to normalising constants, is given by considering the following relationship:

$$\langle s_i s_j \rangle_{Q_1} = Q_1(s_i = 1, s_j = 1) = e^{h_i + h_j + 2J_{ij}} \frac{Z_{-[i,j]}}{Z} \tag{41}$$

where $Z_{-[i,j]}$ is the normalising constant of the original network in which nodes $i$ and $j$ have been removed and the biases are transformed to $h'_k = h_k + J_{ik} + J_{jk}$. This means that we need to evaluate $O(n^2)$ normalising constants to approximate the correlations. To approximate the normalising constants, we use the higher-order expansion (33). Similarly, we attempt to obtain a more accurate approximation to the means $\langle s_i \rangle$, based on the identity,

$$\langle s_i \rangle_{Q_1} = Q_1(s_i = 1) = e^{h_i} \frac{Z_{-[i]}}{Z} \tag{42}$$

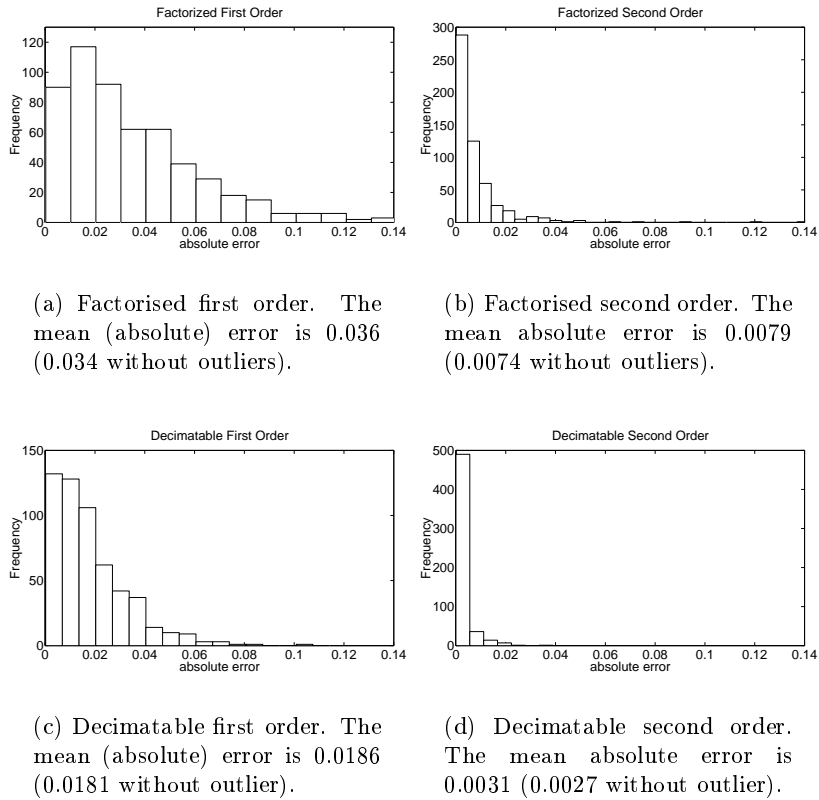

(a) Factorised first order. The mean (absolute) error is 0.036 (0.034 without outliers).

(b) Factorised second order. The mean absolute error is 0.0079 (0.0074 without outliers).

(c) Decimatable first order. The mean (absolute) error is 0.0186 (0.0181 without outlier).

(d) Decimatable second order. The mean absolute error is 0.0031 (0.0027 without outlier).

Figure 3: Approximation of the normalizing constant of a fully connected, eight-node Boltzmann machine (see Figure 2(c)), whose parameters are randomly drawn from a zero-mean, Gaussian distribution with unit variance. A factorised model (see Figure 2(a)) is used as the approximating distribution in (a) and (b) and a decimatable model (see Figure 2(b)) is used in (c) and (d). The histograms are based on 550 different random drawings of the parameters. For the readability of the histogram, we have excluded five outliers (four in (a) and (b), one in (c) and (d)), for which the synchronous mean field iteration did not converge. The mean error is the mean of the absolute value of equation (39).

where $Z_{-[i]}$ is the normalising constant of the original network in which node $i$ has been removed and the biases are transformed to $h'_k = h_k + J_{ik}$. The "intractable" Boltzmann machine in our simulations has 8 fully connected nodes with biases and weights drawn from a standard zero-mean, unit-variance Gaussian distribution. In figures 5 and 6 we plot results for approximating the means $\langle s_i \rangle$ and correlations $\langle s_i s_j \rangle$ respectively. As can be seen, the approach we take to approximate the correlations is an improvement on the standard variational factorised model. We emphasize that, since we are primarily concerned with approximating the normalising constant, there may be more suitable methods dedicated to the task of approximating the correlations themselves. Indeed, one of the drawbacks of these perturbative approaches is that physical constraints, such as that moments should be bounded between 0 and 1, can be violated.

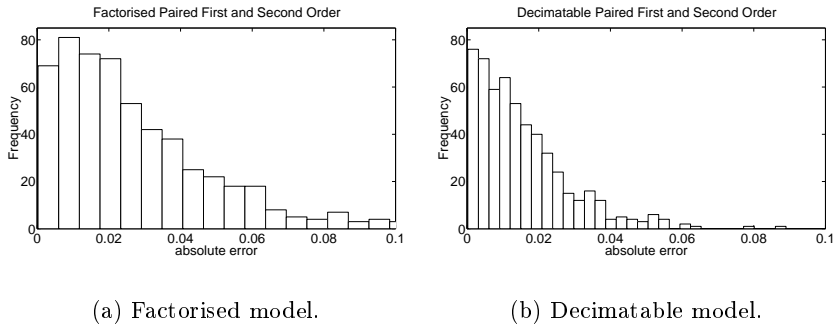

(a) Factorised model.  (b) Decimatable model.

Figure 4: Difference between first and second-order approximation of the normalizing constant of a fully connected, eight-node Boltzmann machine using (a) a factorised and (b) a decimatable model. The histograms are based on the same 550 random drawings as used in Figure 3. All 1100 differences, including the five non-convergent runs, were larger than zero. Again for readability of the histogram, the non-convergent runs are not included. The mean difference was 0.0281 (0.0269 without the non-convergent runs) and 0.0155 (0.0154 without the non-convergent run) for the factorised model and the decimatable model, respectively.

### 5.2.3 LEARNING

The results of the previous section show that, for randomly chosen connection matrices $J$, the higher-order terms can significantly improve the approximation to the normalising constant of the distribution. However, in a learning scenario, such results may be of little interest since the connection matrix will become intimately related to the patterns to be learned—that is, $J$ will take on a form that is appropriate for storing the patterns in the training set as learning progresses.

We are therefore interested here in training networks on a set of visible patterns. That is, we split the nodes of the network into a set of visible $S_V$ and a set of hidden units $S_H$ with the aim of learning a set of patterns $S_V^1 \ldots S_V^p$ with a Boltzmann machine $Q_1(S_V, S_H)$. By using the KL divergence between an approximating distribution $\tilde{Q}_0(S_H|S_V)$ on the hidden units and the conditional distribution $Q_1(S_H|S_V)$ we obtain the following bound

$$\ln Q_1(S_V) \geq -\int \tilde{Q}_0(S_H|S_V) \ln \tilde{Q}_0(S_H|S_V) + \int \tilde{Q}_0(S_H|S_V) \ln Q_1(S_H, S_V) \qquad (43)$$

For the case of Boltzmann machines, $\ln Q_1(S_H, S_V) = H(S_H, S_V) - \ln Z$ in which $\ln Z$ is (assumed) intractable. In order to obtain an approximation to this quantity, we therefore introduce a further variational distribution, $Q_0(S_V, S_H)$ (not the same as $\tilde{Q}_0(S_H|S_V)$) which can be used as in (10) to obtain a lower bound on $\ln Z$. Unfortunately, used in (43), this

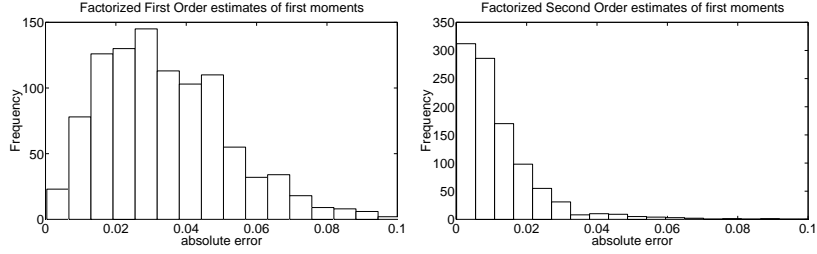

(a) The mean absolute error in approximating the means $\langle s_i \rangle$ is 0.0350.

(b) The mean absolute error in approximating the means $\langle s_i \rangle$ is 0.0129.

Figure 5: Approximating the means $\langle s_i \rangle$ for 1000 randomly chosen 8 node fully connected Boltzmann machines. The weights were drawn from the standard normal distribution. (a) The standard variational approach in which the means are estimated by the means of the best approximating factorised distribution. (b) Using the ratio of normalising constants (41), in which second-order corrections are included in approximating the normalising constants. Using the ratio of normalising constants without the higher order corrections gave a mean error of 0.046, slightly worse than the standard result variational result.

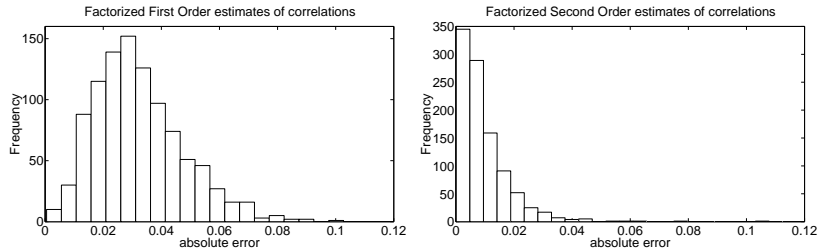

(a) Mean absolute error = 0.0329

(b) Mean absolute error = 0.0103

Figure 6: Approximating the correlations $\langle s_i s_j \rangle$ for 1000 randomly chosen 8 node fully connected Boltzmann machines. The weights were drawn from the standard normal distribution. (a) The standard variational approach in which the means are estimated by the means of the best approximating factorised distribution, and (b) using the ratio of normalising constants, including second-order corrections. Without second-order corrections, this gives a mean error of 0.0428.

lower bound on $\ln Z$ does not give a lower bound on the likelihood of the visible units $Q_1(S_V)$. Nevertheless, we may hope that the approximation to the likelihood gradient is sufficiently accurate such that ascent of the likelihood can be made. Taking the derivative of (43) with respect to the parameters of the Boltzmann machine $Q_1(S_H, S_V)$, we arrive at the following learning rule for gradient ascent given a pattern $S_V$:

$$\Delta J = \eta \left( \langle s_i s_j \rangle_{\tilde{Q}_0(S_H|S_V)} - \langle s_i s_j \rangle_{Q_1(S_H,S_V)} \right) \tag{44}$$

where $\eta$ is a chosen learning rate. The correlations in the first term of (44) are straightforward to compute in the case of using a factorised distribution $\tilde{Q}_0$. The "free" expectations in the second term are more troublesome and need to be approximated in some manner. We examine using the standard factorised model approximations for the free correlations and more accurate "higher-order" approximations as described in section (5.2.2). Here, we are primarily concerned with monitoring the likelihood bound (43) under such gradient dynamics, so we will look at the exact bound value, it's approximation using a factorised model for $\ln Z$ (29), and it's second-order correction (33).

We consider a small learning problem with 3 hidden units and 4 visible units. Ten visible training patterns were formed in which the elements were chosen to be on (value=1) with probability 0.4. In figure 7(a) we demonstrate variational learning in which the free statistics, required for the likelihood gradient ascent rule (44), are approximated with a simple factorised assumption, $\langle s_i s_j \rangle_{Q_1(S_H,S_V)} \approx m_i m_j (i \neq j)$. We display the time evolution of the exact training pattern likelihood bound[3] (43) (solid line), and two approximations to it: the first approximates the intractable $\ln Z$ term by using the standard variational approach with a factorised model (dashed line). The second approximation uses the second-order correction to the factorised model value for $\ln Z$ (dot-dash line). The learning rate was fixed at 0.05. In monitoring the progress of learning, the higher-order correction to the likelihood bound can be seen to be a more accurate approximation of the likelihood compared to that given by the use of the standard variational approximation alone. Interestingly, using the second-order correction to the likelihood, a maximum in the likelihood is detected at a point close to saturation of the exact bound. This is not the case using the first-order approximation alone and, with the standard approximation to the likelihood, the dynamics of the learning process, in terms of the training set likelihood, are poorly monitored.

In figure 7(b) the gradient dynamics are provided again by equation (44) but now with the free correlations $\langle s_i s_j \rangle_{Q_1(S_H,S_V)}$ approximated by the more accurate ratio-of-normalising-constants method, as described in section (5.2.2). Again, we see an improvement in the accuracy of monitoring the progress of the learning dynamics by the simple inclusion of the higher-order term and, again, the higher-order approximation displays a maximum at roughly the correct position. The likelihood is higher than in figure 7(a) since the gradient of the likelihood bound is more accurately approximated through the more accurate correlation estimates. Note that the reason that the exact likelihood lower bound can be above 1 is due to pattern repetitions in the training set. The instabilities in learning after approximately 120 updates are due to the emergence of multimodality in the learned distribution $Q_1(S_H, S_V|J)$, indicating that a more powerful approximation method, able

---

3. In practice, the likelihood on a test set is more appropriate. However, the small number of possible patterns here (16) makes it rather difficult to form a representative test set.

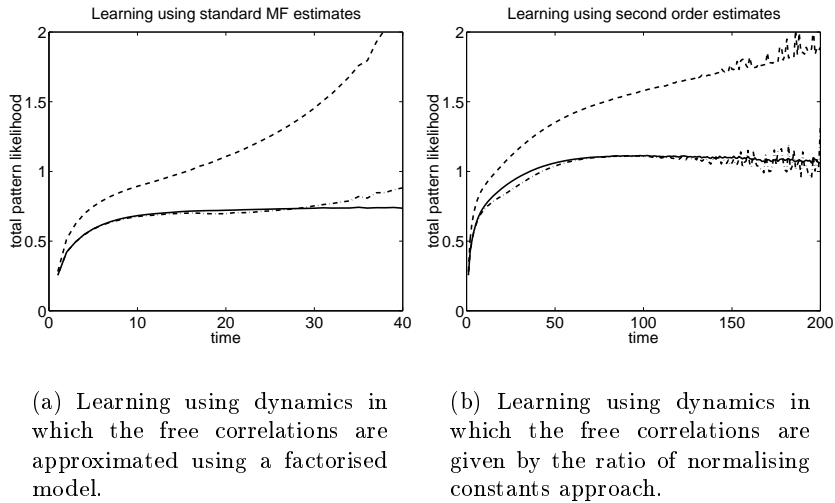

(a) Learning using dynamics in which the free correlations are approximated using a factorised model.

(b) Learning using dynamics in which the free correlations are given by the ratio of normalising constants approach.

Figure 7: Learning a set of 10 random, 4-visible-unit patterns with a Boltzmann machine with 3 hidden units. The solid line is the exact value for the total likelihood bound (43) of all the patterns in the training set. The dashed line is the likelihood bound approximation based on using a factorised model to approximate $\ln Z$, equation (29). The dash-dot line uses the second order correction to $\ln Z$, equation (33).

to capture such multimodal effects, should be used to obtain further improvements in the likelihood.

## 6. Discussion

In this article we have described perturbational approximations of intractable probability distributions. The approximations are based on a Taylor series using a family of probability distributions that interpolate between the intractable distribution and a tractable approximation thereto. These approximations can be seen as an extension of the variational method, although they no longer bound the quantities of interest. We have illustrated our approach both theoretically and by computer simulations for Boltzmann machines. These simulations showed that the approximation can be improved beyond the variational bound by including higher-order terms of the corresponding cumulant expansion.

Simulations showed that the accuracy in monitoring the training set likelihood during the learning process can be improved by including higher order corrections. However, these perturbational approximations cannot be expected to consistently improve on zeroth order (variational) solutions. For instance, if the distribution is strongly multimodal, then using a unimodal variational distribution cannot be expected to be improved much by the inclusion of higher-order perturbational terms. On the other hand, higher-order corrections to multimodal approximations may well improve the solution considerably.

We elucidated the relationship of our approach to TAP, arguing that our approach is expected to offer a more stable solution. Furthermore, the application of our approach to models other than the Boltzmann machine is transparent. Indeed, these techniques are readily applicable to other variational methods in a variety of contexts both for discrete and continuous systems (Barber & Bishop, 1998; Wiegerinck & Barber, 1998).

One drawback of the perturbational approach that we have described is that known "physical" constraints, for example that moments must be bounded in a certain range, are not necessarily adhered to. It would be useful to develop a perturbation method that ensures that solutions are at least physical.

We hope to apply these methods to variational techniques other than the standard Kullback-Leibler bound, and also to study the evaluation of other suitable criteria for the setting of the variational parameters.

## Acknowledgments

We would like to thank Tom Heskes, Bert Kappen, Martijn Leisink, Peter Sollich and Wim Wiegerinck for stimulating and helpful discussions and the referees for excellent comments on an earlier version of this paper.

## Appendix A.

In this appendix we describe how to optimize the variational bound when the normalization constant of an intractable Boltzmann machine is approximated using another tractable Boltzmann machine. For convenience, we denote the potential of an intractable Boltzmann Machine as

$$H_1 \equiv \sum_I w_I s_I \,, \tag{45}$$

where the "extended" parameters are given by $\theta_I \in \{\theta_i, \theta_{ij} | i \in [1, \ldots, N], j = [i+1, \ldots, N]\}$ and $s_I \in \{s_i, s_i s_j | i \in [1, \ldots, N], j = [i+1, \ldots, N]\}$. The potential of the other tractable Boltzmann Machine is denoted as

$$H_0 \equiv \sum_{I \in \Psi} \theta_I s_I \,, \tag{46}$$

where $\Psi$ denotes the set of parameters of the tractable Boltzmann Machine.

We want to optimize the bound of equation (20) with respect to the free parameters $\theta_J$ ($J \in \Psi$), i.e.,

$$\frac{\partial}{\partial \theta_J} [\log Z_0 + \langle \Delta H \rangle_0] = 0 \,, \tag{47}$$

which leads to

$$\langle \Delta H s_J \rangle_0 - \langle \Delta H \rangle_0 \langle s_J \rangle_0 = 0 \tag{48}$$

and thus to the fixed point equation[4]

$$\theta_I = \sum_{J \in \Psi} \sum_K [F_{\Psi\Psi}{}^{-1}]_{IJ} F_{JK} w_K \, , \tag{49}$$

where the Fisher matrix is given by $F_{IJ} = \langle s_I s_J \rangle_0 - \langle s_I \rangle_0 \langle s_J \rangle_0$ which depends on the free parameters $\boldsymbol{\theta}$, and $F_{\Psi\Psi}$ is the Fisher matrix of the tractable Boltzmann machine. For a factorised approximating model, $F_{\Psi\Psi}$ is a diagonal matrix and equation (49) simplifies to equation (32). For a decimatable Boltzmann machine, the elements of the Fisher matrix are tractable. The iteration can be performed either synchronously or asynchronously (Peterson & Anderson, 1987). We prefer here the synchronous case since for all $N$ parameter-updates we only need to calculate and invert a single Fisher matrix instead of $N$ matrices in the asynchronous case. In most applications, however, the asynchronous method seems to be advantageous with respect to convergence (Peterson & Anderson, 1987; Ansari et al., 1995; Saul et al., 1996).

---

4. This fixed point equation is the generalized mean field equation. This form of iteration is not restricted to Boltzmann machines but is a general characteristic of exponential models.